\documentclass[runningheads]{llncs}

\usepackage[T1]{fontenc}
\usepackage{graphicx}
\usepackage{hyperref}
\usepackage{color}

\usepackage{textcomp}
\usepackage{fontawesome}

\begin{document}
\title{EchoCoTr: Estimation of the Left Ventricular Ejection Fraction from Spatiotemporal Echocardiography}
\titlerunning{EchoCoTr}

%
\authorrunning{R. Muhtaseb \& M. Yaqub}

\author{Rand Muhtaseb
\orcidID{0000-0003-2604-5429}
\textsuperscript{(\faEnvelopeO)} \and
Mohammad Yaqub
\orcidID{0000-0001-6896-1105}}
%index{Muhtaseb, Rand}
%index{Yaqub, Mohammad}

\institute{Mohamed Bin Zayed University of Artificial Intelligence, Abu Dhabi, United Arab Emirates\\ \email{\{rand.muhtaseb,mohammad.yaqub\}@mbzuai.ac.ae}}

\maketitle              % typeset the header of the contribution
\begin{abstract}
Learning spatiotemporal features is an important task for efficient video understanding especially in medical images such as echocardiograms. Convolutional neural networks (CNNs) and more recent vision transformers (ViTs) are the most commonly used methods with limitations per each. CNNs are good at capturing local context but fail to learn global information across video frames. On the other hand, vision transformers can incorporate global details and long sequences but are computationally expensive and typically require more data to train. In this paper, we propose a method  that addresses the limitations we typically face when training on medical video data such as echocardiographic scans. The algorithm we propose (EchoCoTr) utilizes the strength of vision transformers and CNNs to tackle the problem of estimating the left ventricular ejection fraction (LVEF) on ultrasound videos. We demonstrate how the proposed method outperforms state-of-the-art work to-date on the EchoNet-Dynamic dataset with MAE of 3.95 and $R^2$ of 0.82. These results show noticeable improvement compared to all published research. In addition, we show extensive ablations and comparisons with several algorithms, including ViT and BERT. The code is available
at \url{https://github.com/BioMedIA-MBZUAI/EchoCoTr}.
\end{abstract}

\section{Introduction}

In medical imaging, there are different imaging modalities that are crucial to real-time clinical assessment and visualization. An example of this is echocardiography, which produces spatiotemporal data made of a sequence of two-dimensional (2D) images. When dealing with spatiotemporal data, it is essential to learn the spatial information as well as take into account the temporal factor in these sequences for an accurate diagnosis. In order to detect abnormalities and certain diseases, cardiologists also tend to take into consideration the temporal information when measuring the left ventricular ejection fraction (LVEF) or while assessing heart wall motion \cite{spatiotemporal_cardiac}. LVEF can be measured as the difference in the left ventricle volume at end-diastole and end-systole divided by the end-diastolic volume estimated from the apical four-chamber (a4c) or apical-two chamber (a2c) views of the heart. LVEF is an important biomarker that can predict heart failure (HF), which is a serious condition that can be caused when the heart cannot pump enough blood and consequently, oxygen to other parts of the body. In 2018, heart failure contributed to 13.4\% of the recorded deaths in the United States \cite{HFstats}. Early diagnosis  of HF  will help cardiologists prescribe medications and encourage patients to have effective lifestyles \cite{Wang}. Heart failure is typically diagnosed if LVEF is less than the normal range (50-80\%). Echocardiography is the most common imaging modality used to assess cardiac function by measuring the left ventricle volume, wall thickness and LVEF since it is real-time, low-cost, ionizing radiation free, portable and a highly sensitive tool compared to other modalities. However, ultrasound technology has many drawbacks, such as operator-dependence, noise, artifacts and decreased contrast that may affect its quality which could lead to a high inter- and intra- observer variability in the diagnosis \cite{cardiacfunction}.

In this paper, we study the impact of different CNNs and transformer models to estimate left ventricle ejection fraction (LVEF) from ultrasound videos. Convolutional neural networks (CNNs) have shown great success when training the models to tackle problems in medical or natural images. However, vision transformers have shown that they may be good contenders to CNNs when solving certain image analysis problems.  There are major differences between the two approaches. CNNs have limited receptive fields in the initial layers, but can progressively enlarge the field of view through convolution operations. In contrast, vision transformers (ViTs), can have the entire field of view starting from the initial layers through the self-attention process. However, unlike CNNs, ViTs do not have inductive bias and hence typically require a large amount of data to train on which is not always available especially in medical imaging. A research study shows that the initial layers of a ViT cannot acquire local information if the dataset is small, which highly impacts the model accuracy \cite{raghu2021vision}. Hence, having a method that combines the strengths of both CNNs and ViTs, to work efficiently with spatiotemporal data in medical imaging assessment, is of great value.

\textbf{Our contribution} in this work is three fold: 
\begin{itemize}
    \item We propose EchoCoTr (\textbf{Echo Co}nvolutional \textbf{Tr}ansformer) which is a method that is able to analyze echocardiography video sequences by combining the strength of CNNs and vision transformers to accurately estimate the heart's ejection fraction. Even though EchoCoTr is adapted from UniFormer \cite{uniformer} which worked on natural video datasets, some changes were made to address the challenging problems we face such as proper frame sampling. 
    \item We show how our proposed method outperforms all published work to-date on a large scale public dataset \cite{echonet,uvt}, which does not require: 1) information regarding the position of end-systolic (ES) and end-diastolic (ED) frames, 2) segmentation masks as EchoNet-Dynamic’s beat-to-beat pipeline \cite{echonet}, and 3) a pre-defined length of the cardiac scan. 
    \item We compare our proposed method with several existing deep learning algorithms and perform thorough ablation studies to provide a deep discussion of the results.
\end{itemize}

\section{Related Works}
Many research papers \cite{Zhang,Smistad,Smistad2,contrastive} were introduced to improve the segmentation of the left ventricle to accurately estimate ejection fraction. Silva et al. \cite{Silva} used a 3D CNN with residual learning blocks to estimate ejection fraction from transthoracic echocardiography (TTE) exams. Ouyang et al. \cite{echonet} proposed a deep learning approach to estimate the beat-to-beat ejection fraction and predict heart failure with reduced ejection fraction (HFrEF) by combining the semantic segmentation results and the clip-level ejection fraction prediction using spatiotemporal CNN \cite{r2plus1d}. Recently, Reynaud et al. \cite{uvt} proposed a transformer model based on  residual auto-encoder to reduce the dimensions followed by Bidirectional Encoder Representations from Transformers (BERT) for end-systolic (ES) and end-diastolic (ED) frame detection and ejection fraction estimation. Understanding spatiotemporal data using transformers can also be found in other medical imaging domains. Latest research areas have been focusing on using transformers to diagnose COVID-19 \cite{hsu2021visual,ZhangLei,ZhangLei2} and perform 3D image segmentation of multi-organ and on brain tumor datasets \cite{hatamizadeh2021unetr}.

A recent work was proposed by Li. et al \cite{uniformer} in a modified transformer version that combines the strengths of 3D CNNs and spatiotemporal transformers. The UniFormer has three main components. The first component is Dynamic Position Embeddings (DPE) which maintains the spatiotemporal positions of the video tokens by applying 3D depthwise convolution without padding. The second component is Multi-Head Relation Aggregator (MHRA) which learns the local token relations to ignore the redundancy due to the small differences found in adjacent frames in the initial layers. However, in the last two stages, MHRA learns the global token affinity, which is similar to the self-attention scheme. The last component is Feed Forward Network (FFN) which has two linear layers.

\section{Methods}

\begin{figure}[t!]
    \centering
    \includegraphics[scale=0.42]{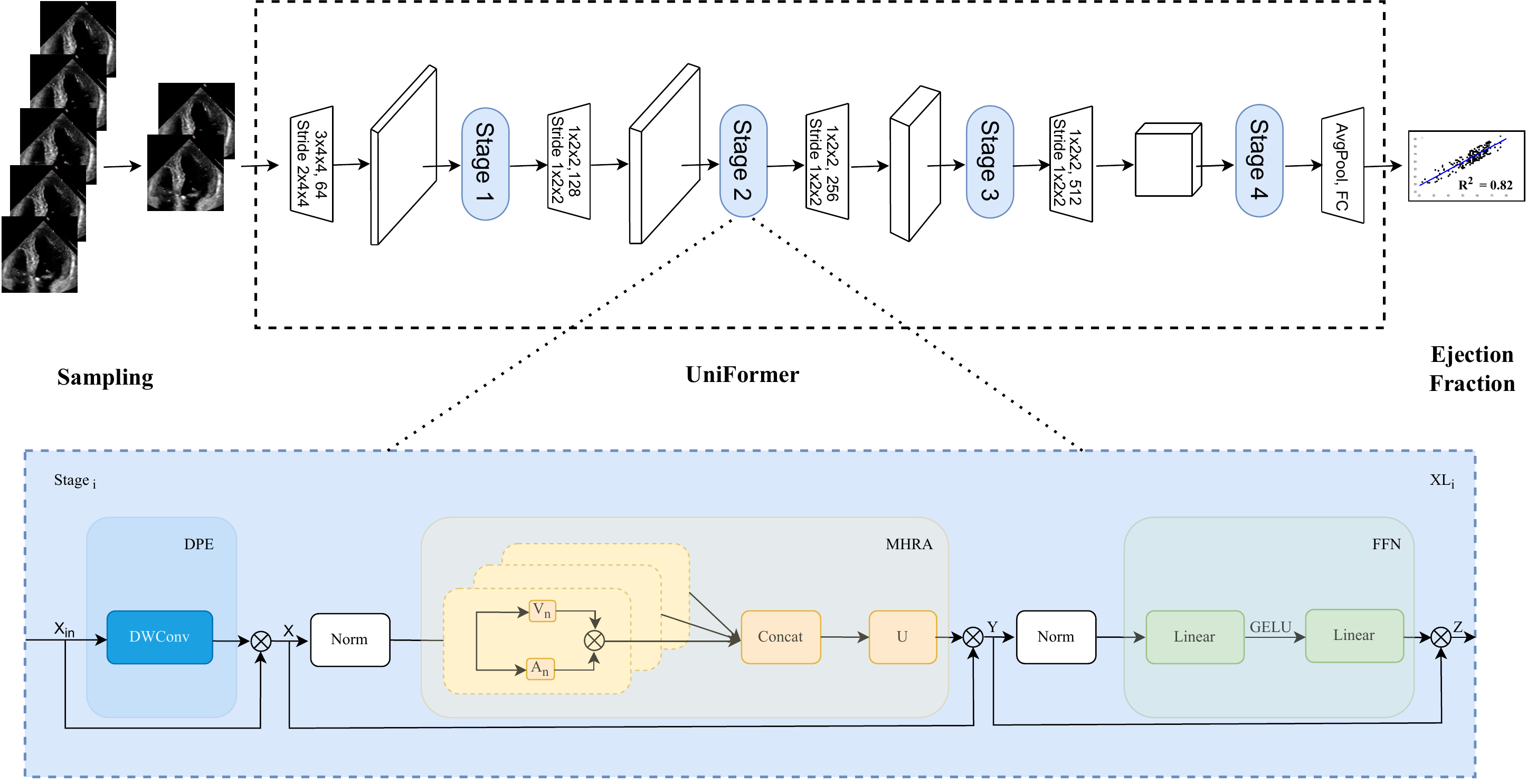}
    \caption{The overall architecture of EchoCoTr is based on UniFormer \cite{uniformer}. Echocardiographic videos will be first sampled, to introduce dissimilarity between the frames, then fed to the UniFormer model to predict the LVEF for the entire video sequence.}
    \label{fig:arch}
\end{figure}

In this section, we describe the frames sampling approach, model architecture and the proposed method when estimating ejection fraction from echocardiographic videos.

\subsection{Frames Sampling} \label{sampling}
Deep learning networks require a fixed number of video frames from each scan. However, EchoNet-Dynamic videos contain one or more cardiac cycles, which also vary in the number of frames per cycle  (approximately 20-30 frames). Moreover, the differences between the adjacent video frames are small. Because of that, we had to perform a video frame sampling by experimenting with different number of frames \{32, 36, 40\} and uniform frequencies \{2, 4, 6\} adapted by \cite{echonet}. The sampling operation starts with a random clip within the range of [0 - (Number of original video frames - (Number of sampling video frames - 1) * Sampling frequency)]. Prior to that, in the case of short videos, frames filled with zeros will be added to the end of the video. The strength of using video sampling techniques replaces the traditional methods that clinicians do, which requires knowing the location of ES and ED frames before calculating the LVEF. In addition to that, as the location of ES and ED frames are already known beforehand, we also experimented with only selecting ES and ED frames from the video to check if these are sufficient to give an adequate LVEF prediction. A summary of some experiments related to video sampling is found in \autoref{table:ablation}.

\subsection{Architecture Overview}
EchoCoTr builds on UniFormer \cite{uniformer} to address both the challenges of the local redundant features and the complex dependency among the video frames in the cardiac echo scans. Subtle differences between adjacent frames make it important that the network selects the most representative frames when estimating LVEF. Therefore, we had to adapt an architecture that effectively learns the local features without redundancy in the adjacent frames while capturing the global information along the video. An illustration of the overall architecture is found in Fig. \ref{fig:arch}. Before feeding the ultrasound videos to the UniFormer model to generate LVEF prediction, we sample the video frames to introduce dissimilarity and make sure that there is no redundancy between the neighboring frames. Before each stage in the UniFormer model, 1x2x2 convolution with stride of 1x2x2 is applied. However, to downsample the spatiotemporal dimensions of the input video, 3x4x4 convolution with stride of 2x4x4 is used instead in the first stage. As a method for echocardiography, we experimented with two different UniFormer variants: UniFormer-S and UniFormer-B with the aim of investigating the impact of the number of UniFormer blocks on the LVEF estimation. The number of UniFormer blocks used for EchoCoTr-S (small model) and EchoCoTr-B (baseline version) are \{3, 4, 8, 3\} and \{5, 8, 20, 7\}, respectively. The drop rates are set to 0.1 for EchoCoTr-S and 0.3 for EchoCoTr-B.

\subsection{Existing Methods for LVEF Estimation}
In this subsection, for the sake of ablations and comparisons, we present recent published methods that addressed LVEF prediction. The work of \cite{uvt} has shown that using a BERT model could be used to estimate LVEF. First, the dimensions of the input videos are reduced to a vector of size (Batch Size x Number of Frames) x 1024 using a ResNetAE \cite{ResNetAE} encoder. Two sampling strategies were introduced by \cite{uvt}. The first is mirroring (M), which places the repeated sequence between the ES and ED frames after the last annotated frame. The second strategy is random sampling (R), which adds up 10-70\% of the distance between the two annotated frames before and after the sampled frames from a heart cycle. However, the result that was reported did not outperform \cite{echonet} that used a spatiotemporal convolution based ResNet (ResNet (2+1)D) \cite{r2plus1d}. Therefore, we compare our proposed method with the BERT method \cite{uvt} and with other transformer models, such as DistilBERT and ViT.

\section{Experiments}
In this section, we aim to give a brief summary of the dataset used and experimental setup that we had for our experiments.

\subsection{Datasets}
\textbf{EchoNet-Dynamic} \cite{echonetDataset} is the largest publicly available dataset of echocardiographic scans for the apical four-chamber (a4c) view of the heart acquired from the Stanford University Hospital. It consists of 10,030 videos in total. Each video consists of a sequence of 112x112 grayscale images and traces for the left ventricle end-systole (ES) and end-systole (ED) frames. In addition, every video is labelled with the corresponding end-systolic volume (ESV), end-diastolic volume (EDV) and ejection fraction (EF).

\subsection{Experimental Setup}
The data split sizes for training, validation and testing are 7460, 1288 and 1277, respectively. This is the same split chosen by \cite{echonetDataset}. All selected hyperparameters are optimized experimentally. The evaluation metrics used are mean absolute error (MAE), root mean squared error (RMSE) and R-squared ($R^2$). In addition to that, we also compare the floating point operations (FLOPs) values for the different models using fvcore package \cite{fvcore}.
\\~\\
\textbf{EchoCoTr Experiments}:
EchoCoTr models are trained on an NVIDIA A100 GPU for 45 epochs. The batch sizes used for EchoCoTr-S and EchoCoTr-B are 25 and 16, respectively. AdamW is used as an optimizer with a value of 1e-4 for both the learning rate and weight decay. Both models were pretrained on the Kinetics-400 dataset with different pretraining strategies. EchoCoTr-S is pretrained on 16x1x4 frames with sampling stride of 8. However, the weights used for EchoCoTr-B is 32x1x4 frames with sampling stride of 4. Frame resolutions are kept as same as in the original public dataset (112x112).
\\~\\
\textbf{Other Experiments}
BERT, DistilBERT and ViT models are trained for 5 epochs with batch size of 2, which is small because of the large model size. AdamW is used as an optimizer with a learning rate of 1e-5 and weight decay of 1e-2.  Images are padded to be 128x128 in size to facilitate fair comparison and easy integration for the three models. The Hugging Face Python library is used for the transformer experiments.

\begin{table}
\caption{Comparison with the state-of-the-art results on EchoNet-Dynamic dataset. "R." and "M" are the sampling methods proposed by \cite{uvt}, which refer to Random and Mirroring sampling. EchoNet-Dynamic (1) predicts the clip-level LVEF using 32 frames. EchoNet-Dynamic (2) uses the segmentation and clip-level LVEF outputs to evaluate the beat-to-beat LVEF estimation for the entire video sequence. One sample from the testing dataset is used to calculate the FLOPs.}
\label{table:results}
\centering
\begin{tabular}{|l|c|c|c|c|c|}
\hline
\textbf{Model} & \textbf{No. of Frames} & \textbf{FLOPs} & \textbf{MAE} \textdownarrow & \textbf{RMSE} \textdownarrow & \textbf{R}$^2$ \textuparrow\\
\hline
    UVT R. \cite{uvt} & 128 & 130.00G & 6.77 & 8.70 & 0.48\\
  \hline
  UVT M. \cite{uvt} & 128 & 130.00G & 5.95 &
  8.38 & 0.52\\
  \hline
  R3D \cite{echonet} & 32 & 92.273G & 4.22 & 5.62 & 0.79\\ 
  \hline
  MC3 \cite{echonet} & 32 & 97.656G & 4.54 & 5.97 & 0.77\\
  \hline
  EchoNet-Dynamic \cite{echonet} (1) & 32 & 91.974G & 4.22 &  5.56 &  0.79\\ 
  \hline
  EchoNet-Dynamic \cite{echonet} (2) & beat-to-beat & - & 4.05  & 5.32 & 0.81\\
  \hline
  EchoCoTr-B & 36 & 44.907G & \textbf{3.98} & \textbf{5.34} & \textbf{0.81}\\ 
  \hline
  EchoCoTr-S & 36 & 19.611G & \textbf{3.95} & \textbf{5.17} &  \textbf{0.82}\\ 
  \hline
\end{tabular}
\end{table}

\section{Results}

As \autoref{table:results} shows, our EchoCoTr-S model, which was trained on only 36 frames with sampling frequency of 4 (3.95 MAE), outperforms the state-of-the-art results reported by \cite{echonet,uvt}. It is also noticeable from the results that the EchoCoTr-S experiment (3.95 MAE) performed slightly better than EchoCoTr-B (3.98 MAE). We test the effect of various sampling frequencies and sizes on the LVEF prediction. Results in \autoref{table:ablation} show that a sampling frequency of 4 frames achieves the best result for both small and baseline models. In addition, the optimal number of frames is found to be 36 for both models. Surprisingly, training both EchoCoTr-S and EchoCoTr-B models on only two frames (ES and ED) from each video achieves lower yet satisfactory results (4.432 and 4.494 MAE).
\\~\\
\autoref{table:ablation} also displays the results of our experiments that we performed using BERT, DistilBERT and ViT. We only report the experiments for the mirroring sampling strategy, as it achieved better results than the random one in \cite{uvt}. Results suggest that the BERT model with the mirroring sampling on 36 and 128 frames (5.788 and 5.950 MAE, respectively) \cite{uvt} performs better than DistilBERT and ViT when estimating LVEF. Moreover, reducing the number of frames to 36 was negatively impacting 
DistilBERT's MAE score the most (6.689).

\begin{table}
\caption{\textbf{Ablation study}: Summary of experiments performed on the EchoNet-Dynamic Dataset using EchoCoTr and transformer models. The sampling strategy used for BERT, DistilBERT and ViT experiments is mirroring \cite{uvt}. $2^*$ refers to the two video frames used, which are ES and ED.}\label{table:ablation}
\centering
\begin{tabular}{|l|c|c|c|c|c|c|c|}
\hline \textbf{Model} & \textbf{Frequency} & \textbf{No. of Frames} & \textbf{Batch Size} & \textbf{MAE} \textdownarrow & \textbf{RMSE} \textdownarrow & \textbf{R}$^2$ \textuparrow\\
\hline
BERT & - & 36 & 2 & 5.788 & 8.137 & 0.545\\
\hline
BERT \cite{uvt} & - & 128 & 2 & 5.950 & 8.380 & 0.520\\
\hline
DistilBERT & - & 36 & 2 & 6.689 & 9.234 & 0.414\\
\hline
DistilBERT & - & 128 & 2 & 6.430 & 8.940 & 0.451\\
\hline
ViT & - & 36 & 2 & 6.454 & 8.955 & 0.448\\
\hline
ViT & - & 128 & 2 & 6.527 & 9.053 & 0.436\\
\hline
% ViT & R. & 128 & 2 & 6.148 & 8.282 & 0.529\\
% \hline
EchoCoTr-S & - &  $2^{*}$  &  25  & 4.432 & 5.998 & 0.759\\
  \hline
EchoCoTr-S & 2 &  36  &  25  & 4.168 & 5.541 & 0.795 \\
  \hline
EchoCoTr-S & 4 &  32  &  25  & 3.966 & 5.290 & 0.813 \\
  \hline
EchoCoTr-S & 4 &  36  &  25  & \textbf{3.947} & \textbf{5.174} & \textbf{0.821} \\
  \hline
EchoCoTr-S & 4 &  40  &  25  & 4.010 & 5.326 & 0.810 \\
  \hline
EchoCoTr-S & 6 &  36  &  25  & 4.135 & 5.434 & 0.803 \\
  \hline
EchoCoTr-B & - &  $2^{*}$  &  16  & 4.494 & 6.205 & 0.743\\
  \hline
EchoCoTr-B & 2 &  36  &  16  & 4.184 & 5.590 & 0.791\\
  \hline
EchoCoTr-B & 4 &  36  &  16  & \textbf{ 3.980} &\textbf{ 5.342} & \textbf{0.809}\\
  \hline
EchoCoTr-B & 6 &  36  &  16  & 4.068 & 5.410 & 0.804\\
  \hline
\end{tabular}
\end{table}

\section{Discussion}
In this paper, we propose EchoCoTr which is a method that combines the strengths of 3D CNNs and vision transformers for spatiotemporal echocardiography assessment in order to estimate LVEF on ultrasound videos.
\\~\\
The results in \autoref{table:results} show that the model trained using EchoCoTr-S on only 36 frames with a uniform sampling frequency of 4 (3.95 MAE), outperforms the state-of-the-art results reported by EchoNet-Dynamic on the beat-to-beat pipeline for LVEF prediction (4.05 MAE). In addition, unlike EchoNet-Dynamic, our method does not require the segmentation masks. Furthermore, our score is also better than the result that EchoNet-Dynamic stated for 32 frames and a sampling frequency of 2 frames (4.22 MAE). As illustrated in \autoref{table:ablation}, a proper video sampling strategy plays a role in improving the results when using EchoCoTr models. This might be due to the different details that the model attends to spatially and temporally. For instance, not all adjacent frames might be needed during training and frames from multiple heart cycles are likely needed to provide a better temporal representation. Furthermore, we think that 36 frames with sampling frequency of 4 is found to be an ideal configuration to the problem at hand, because it covered multiple cardiac cycles (4-5 cycles) while skipping redundant and similar frames in most of the videos found in the EchoNet-Dynamic dataset. Hence, this has led to a more accurate estimation of LVEF prediction for the entire video. In fact, the frame sampling strategy we propose is aligned with the clinical guidelines that suggest estimating LVEF from up to 5 cardiac cycles.
\\~\\
Another remarkable result found is that EchoCoTr achieves satisfactory LVEF estimations when trained on only two frames (ES and ED). Due to its design, it ignores the local redundant features but learns the long-range dependencies. This follows the same methodology that clinicians do when calculating the EDV and ESV values to estimate LVEF. 
\\~\\
It is also clearly seen from \autoref{table:ablation} that training on 36 frames achieves comparable results to 128 frames for BERT, DistilBERT and ViT models. However, all these experiments did not perform as well as our proposed method. We hypothesize that these models could not  capture the temporal information as effectively as our proposed method while learning the local features within different frames. We believe that EchoCoTr-B performed marginally less than EchoCoTr-S due to its large architectural size that might be an overkill for the LVEF estimation problem.

\section{Conclusion}
We propose EchoCoTr which utilizes CNNs' discriminative spatial ability with transformers' temporal perception to estimate LVEF from a set of sampled frames from multiple heart cycles. The method outperforms other recent work when estimating ejection fraction on the EchoNet-Dynamic dataset. The goal of this paper is not to comprehensively study the performance of different transformer models, but to compare their performances with our CNN-Transformer method on spatiotemporal image analysis.
For future work, it is valuable to study the effect of self-supervision on EchoCoTr's performance by using the unlabelled frames from each video. EchoNet-Dynamic dataset size proved to be enough to produce good results using EchoCoTr and spatiotemporal convolutional neural networks. Furthermore, it is also worth experimenting with the impact of performance on smaller datasets and datasets with abnormal motion of the heart.

\section{Acknowledgments}
We thank Mohamed Bin Zayed University of Artificial Intelligence (MBZUAI) for providing funding for this study, and Mohamed Saeed for providing his support.
\bibliographystyle{splncs04}
\bibliography{paper2304}

\end{document}